\documentclass{article}
\usepackage{spconf,amsmath,graphicx}

\usepackage{enumitem}
\usepackage{tabularx}
\usepackage{float}
\usepackage{booktabs}
\usepackage[table]{xcolor} 
\usepackage{makecell}
\usepackage{siunitx}
\usepackage{subcaption}
\usepackage{xcolor}
\usepackage{graphicx}
\usepackage{multirow}
\usepackage{hyperref}
\hypersetup{
    colorlinks,
    linkcolor={blue!50!black},
    citecolor={blue!50!black},
    urlcolor={blue!50!black}
}
\usepackage{svg}

\definecolor{darkblue}{RGB}{0, 76, 153}
\definecolor{darkgreen}{RGB}{0, 153, 76}

\newcolumntype{Y}{>{\centering\arraybackslash}X}
\newcolumntype{C}{>{\centering\arraybackslash}m{1.75cm}} 
\newcolumntype{Z}{>{\centering\arraybackslash}m{10cm}} 

\setlist{nosep, leftmargin=14pt}

\usepackage{mwe} 

\title{Pixel-level Counterfactual Contrastive Learning\\for Medical Image Segmentation}
\name{Marceau Lafargue-Hauret, Raghav Mehta, Fabio De Sousa Ribeiro, Mélanie Roschewitz, Ben Glocker}
\address{Department of Computing, Imperial College London, UK}
\begin{document}
%
\maketitle
\begin{abstract}
Image segmentation relies on large annotated datasets, which are expensive and slow to produce. \textit{Silver-standard (AI-generated)} labels are easier to obtain, but they risk introducing bias. Self-supervised learning, needing only images, has become key for pre-training. Recent work combining contrastive learning with counterfactual generation improves representation learning for classification but does not readily extend to pixel-level tasks. We propose a pipeline combining counterfactual generation with dense contrastive learning via Dual-View (DVD-CL) and Multi-View (MVD-CL) methods, along with supervised variants that utilize available silver-standard annotations. A new visualisation algorithm, the Color-coded High Resolution Overlay map (CHRO-map)
is also introduced. Experiments show annotation-free DVD-CL outperforms other dense contrastive learning methods, while supervised variants using silver-standard labels outperform training on the silver-standard labeled data directly, achieving $\sim$94\% DSC on challenging data. These results highlight that pixel-level contrastive learning, enhanced by counterfactuals and silver-standard annotations, improves robustness to acquisition and pathological variations.
\end{abstract}
\section{Introduction}
\label{sec:intro}

Contrastive learning~\cite{sohnImprovedDeepMetric2016,oordRepresentationLearningContrastive2019,chenSimpleFrameworkContrastive2020,heMomentumContrastUnsupervised2020} has become a popular solution to exploit unlabeled data. It aims to learn representations by bringing similar samples (positives) closer in the embedding space while pushing dissimilar samples (negatives) apart. Typically, this is achieved by defining pairs or sets of augmented views of the same image as positives, and different images as negatives, and training the model with a contrastive loss such as the InfoNCE or NT-Xent objective~\cite{sohnImprovedDeepMetric2016,chenSimpleFrameworkContrastive2020}. This framework encourages the encoder to extract features that are invariant to augmentations and discriminative across samples, enabling effective representation learning without labels. 

Classical contrastive methods, such as SimCLR~\cite{chenSimpleFrameworkContrastive2020} or MoCo~\cite{heMomentumContrastUnsupervised2020}, operate at the image level, producing a single embedding per image. While this is effective for classification or retrieval tasks, it limits their ability to capture spatially localized information required for dense prediction tasks such as segmentation. To address this limitation, dense contrastive learning~\cite{o.pinheiroUnsupervisedLearningDense2020,wangDenseContrastiveLearning2021,xiePropagateYourselfExploring2021} extends the contrastive objective to the pixel or patch level. Instead of comparing global image embeddings, pixel-level embeddings are contrasted across multiple augmented views of the same image, enabling spatially consistent and semantically rich representations that can be leveraged for segmentation or detection tasks. Contrastive learning has also been used in supervised or semi-supervised settings~\cite{sohnImprovedDeepMetric2016,huSemisupervisedContrastiveLearning2021}. However, the standard data augmentation used in contrastive learning methods (such as rotation, cropping, and blurring) does not capture the complex and challenging dataset shifts that arise in real-world applications. 

\begin{figure*}[t]
    \centering
    \includegraphics[width=0.9\linewidth]{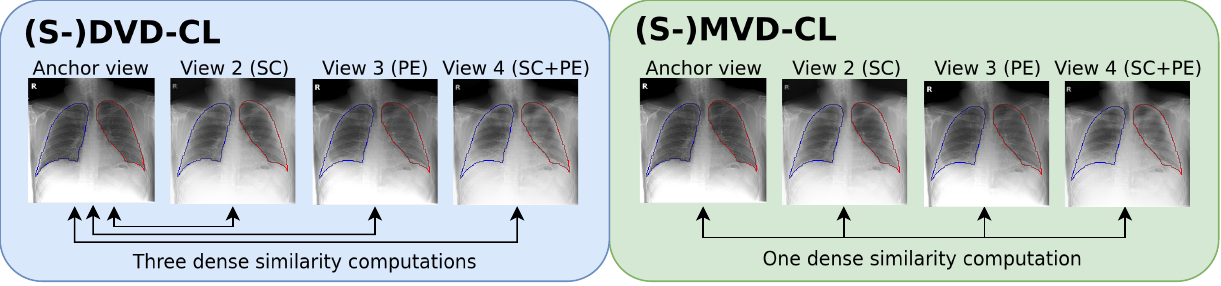}
    \caption{Overview of the proposed methods. Views are formed through scanner (SC) and pleural effusion (PE) counterfactuals, in combination with a traditional augmentation pipeline. (S-)DVD-CL computes three dense similarity computations between the anchor view and each of the target views, and average results. (S-)MVD-CL computes similarity between all views at once.}
    \label{fig:architecture}
\end{figure*}

This limitation could be overcome by counterfactual generation, which seeks to produce realistic and semantically meaningful variations in the data. Counterfactual generation relies on the framework of Structural Causal Models (SCMs), which describe a system as a set of causal variables connected by deterministic functions and exogenous noise terms. By intervening on a specific variable (e.g., changing the presence of pleural in chest x-ray) while keeping other factors fixed, one can generate a counterfactual version of the data that reflects \textit{``what if?"} that change had occurred. Deep SCMs (DSCMs)~\cite{pawlowski2020deep,ribeiroHighFidelityImage2023} extend this idea by parameterizing causal mechanisms with neural networks, enabling counterfactual image synthesis directly in high-dimensional spaces.  The development of counterfactual generation~\cite{pawlowski2020deep,kocaogluCausalGANLearningCausal2018,ribeiroHighFidelityImage2023} has found use in improving model robustness~\cite{chenGeneralizableSingleSourceCrossModality2025,mehtaCFSegCounterfactualsMeet2025} as well as in contrastive learning~\cite{roschewitz2025robust} in image classification models. 

In this work, we leverage counterfactual generation to enhance contrastive learning for \textit{dense} contrastive objectives, marking the first exploration of this setting. We focus on lung segmentation from Chest X-rays in the presence of Pleural Effusion (PE). PE renders a large section of the lungs opaque, often leading to undersegmentation of the lungs~\cite{mehtaCFSegCounterfactualsMeet2025}.
%
Our contributions can be summarized as follows:
\begin{itemize}
    \item Four novel dense-contrastive learning methods: Dual-View (DVD-CL) and Multi-View (MVD-CL), plus their \textit{silver-standard} label supervised variants.
    \item Integration of counterfactual augmentation for invariant representation learning.
    \item CHRO-map for visualising pixel-level embeddings.
\end{itemize}

These innovations enhance the interpretability and robustness of medical image segmentation using unlabeled data and may provide components for constructing foundational models.

\section{Methods}
\label{sec:methods}

In contrastive learning, data augmentations play a crucial role in making models robust to dataset shifts that arise in real-world applications. However, such shifts are often complex and difficult to capture using standard augmentations, such as color jittering or random cropping. To address this limitation, counterfactual generation aims to simulate realistic and semantically meaningful variations in the data.

\subsection{Counterfactual Generation}
We train a counterfactual generative Hierarchical Variational AutoEncoder (HVAE)~\cite{ribeiroHighFidelityImage2023,monteiroMeasuringAxiomaticSoundness2023} using the same method as the authors. We use sex, scanner type, and the presence of pleural effusion (PE) as parents in the causal graph of SCM for Chest X-Ray. We generate three counterfactuals: (i) only the scanner is changed, (ii) only the presence of the PE is changed, and (iii) both the scanner and the presence of PE are changed. Each of these counterfacutals and the base image is further augmented with random rotation, cropping, and image effects (color jittering, Gaussian blurring, and solarization).

\subsection{Dense Counterfactual Contrastive Learning}
We propose four different dense contrastive learning methods that leverage image counterfactuals for data augmentation (see Fig.~\ref{fig:architecture}). They consist of dual- and multi-view approaches, each available in both supervised and unsupervised variants.\\

\noindent\textbf{Dual-View Dense Contrastive Learning (DVD-CL)} extends the pixel-level contrastive framework of VADeR \cite{o.pinheiroUnsupervisedLearningDense2020} to support multiple views. In VADeR, two augmented views of an image are used: pixels at the same spatial location form positive pairs, while all other pixel combinations serve as negatives. In DVD-CL, the non-counterfactual view 
serves as the anchor view. The remaining views, counterfactual images generated by the HVAE, serve as target views.
For each target view, positive pixel pairs are defined as those sharing the same spatial location in the anchor and target view, while all other pixel combinations are negatives. The NT-Xent loss is averaged over both directions, first treating the anchor as the reference, then the target. The overall contrastive loss is calculated by averaging across all target views (three in our experiments, although any number of views are supported). \\

\noindent\textbf{Multi-View Dense Contrastive Learning (MVD-CL)} considers all views simultaneously and forms pixel pairs following the same procedure as in DVD-CL, but extended across all available views. The contrastive loss is then computed over all pairwise pixel similarities, enabling joint optimization across the entire multi-view set. \\

\noindent\textbf{Supervised DVD-CL (S-DVD-CL)} draws inspiration from prior pixel-level contrastive learning methods that leverage annotations to define positive and negative pairs~\cite{huSemisupervisedContrastiveLearning2021,deyLearningGeneralpurposeBiomedical2024}. In this setting, pixels belonging to the same class are treated as positives (ex., left lung), while pixels from different classes are treated as negatives (ex., right lung). Using this supervision-based pairing strategy, we compute the contrastive loss in the same manner as in DVD-CL, evaluating the loss over two views at a time and averaging the results across all view pairs. To ensure meaningful supervision, only non-background pixels are used as anchors. \\

\noindent\textbf{Supervised MVD-CL (S-MVD-CL)} again considers all views simultaneously, forming pixel pairs with annotations in the same manner as in S-DVD-CL. The contrastive loss is computed over all pixel similarities and averaged across the full set of views.

\begin{figure*}[t]
    \centering
    \hfill
    \begin{subfigure}[b]{0.235\textwidth}
        \centering
        \includegraphics[width=0.48\textwidth]{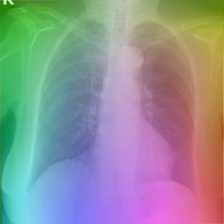}\hspace{2px}%
        \includegraphics[width=0.48\textwidth]{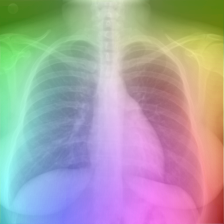}
        \caption{DVD-CL}
        \label{fig:dvd-chromap}
    \end{subfigure}
    \hfill
    \begin{subfigure}[b]{0.235\textwidth}
        \centering
        \includegraphics[width=0.48\textwidth]{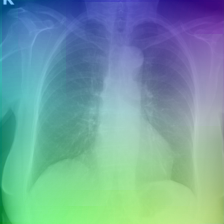}\hspace{2px}%
        \includegraphics[width=0.48\textwidth]{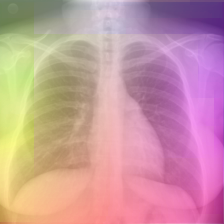}
        \caption{MVD-CL}
        \label{fig:mvd-chromap}
    \end{subfigure}
    \hfill
    \begin{subfigure}[b]{0.235\textwidth}
        \centering
        \includegraphics[width=0.48\textwidth]{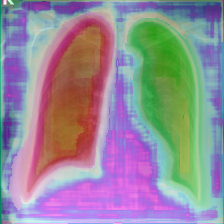}\hspace{2px}%
        \includegraphics[width=0.48\textwidth]{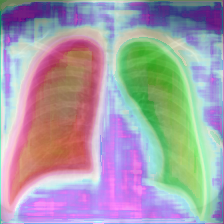}
        \caption{S-DVD-CL}
        \label{fig:sdvd-chromap}
    \end{subfigure}
    \hfill
    \begin{subfigure}[b]{0.235\textwidth}
        \centering
        \includegraphics[width=0.48\textwidth]{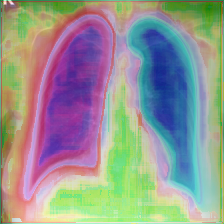}\hspace{2px}%
        \includegraphics[width=0.48\textwidth]{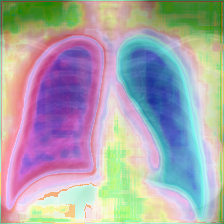}
        \caption{S-MVD-CL}
        \label{fig:smvd-chromap}
    \end{subfigure}\hfill\hfill
    \caption{Output CHRO-maps of the four different methods. Similar colors indicate similar encodings. We observe that MVD-CL fails to capture meaningful representations, whereas DVD-CL effectively encodes pixels based on their relative position to the spine, which can be further fine-tuned for segmentation. S-DVD-CL and S-MVD-CL manage to sharply distinguish both lungs.}
    \label{fig:chro-maps}
\end{figure*}

\subsection{CHRO-map}
We introduce a Color-coded High-Resolution Overlay mapping (CHRO-map) to visualize the network's output embeddings. First, the high-dimensional embeddings are projected to a two-dimensional space using UMAP~\cite{mcinnesUMAPUniformManifold2020}. We then compute the minimum volume enclosing ellipse around the projected embeddings and determine the affine transformation that maps this ellipse to the unit circle. This transformation is applied to all embeddings, which are subsequently assigned HSV colors based on their polar coordinates, using the angle $\theta$ as the hue and the radius $r$ as the value. Finally, each color is overlaid on its corresponding pixel in the image to produce the visualization. CHRO-map provides an interpretable, high-resolution 2D visualization of dense feature embeddings, where semantic proximity in the latent space corresponds to color similarity in the visualization. This enables qualitative assessment of representation clustering and counterfactual invariance (Fig.~\ref{fig:chro-maps}).

\section{Implementation Details}

\noindent\textbf{Dataset.} We utilize the publicly available PadChest~\cite{bustosPadChestLargeChest2020} dataset for experimentation ($\sim$60k training and $\sim$17k validation images). We use \textit{silver-standard} CheXmask~\cite{gaggionCheXmaskLargescaleDataset2024} annotations for the supervised contrastive learning variants. The goal is to generate (left and right) lung segmentation from Chest X-ray images of healthy patients and patients with Pleural Effusion (PE). Furthermore, we utilize 70 manually annotated images (20 healthy and 50 PE) for fine-tuning and validating segmentation models~\cite{mehtaCFSegCounterfactualsMeet2025}.

\noindent\textbf{Network}. Instead of pre-training only the encoder of a U-Net model, as is done in other contrastive learning works~\cite{xiePropagateYourselfExploring2021,wangDenseContrastiveLearning2021}, we pre-train the full encoder-decoder structure of a standard U-Net model with a ResNet50 encoder. 

\noindent\textbf{Precise Pixel Tracking.} We model each geometric transformation (e.g., rotation, cropping, scaling) using a homogeneous transformation matrix. This enables us to maintain exact pixel-level correspondences across augmented views, allowing flexible formation of pixel-level positive pairs under any affine transformation.

\noindent\textbf{Sampling.} As computing the contrastive losses over all pixels is computationally infeasible, we randomly sample 1,000 pixels from each view. Sampling is restricted to the intersection of active views, specifically the anchor and target views for (S-)DVD-CL, and all views for (S-)MVD-CL. 
This sampling strategy is particularly suitable when the objects to be segmented are located near the center of the image, as pixels closer to the borders are less likely to be observed during pre-training.

\section{Results}
\label{sec:results}


\begin{table}[t]
\centering
\caption{Evaluation of the learned embedding using the proposed contrastive learning methods before fine-tuning.}
\begin{tabular}{>{\centering\arraybackslash}p{65pt} 
                >{\centering\arraybackslash}p{65pt} 
                >{\centering\arraybackslash}p{80pt}}
\toprule
         & d/$\sigma$ ($\uparrow$)       & K-means-purity ($\uparrow$) \\ \midrule
DVD-CL   & 0.83                          & 0.6979 \cellcolor{darkgreen!10} \\
MVD-CL   & 1.39 \cellcolor{darkgreen!10}                         & 0.6391 \\ \midrule
S-DVD-CL & \textbf{17.07} \cellcolor{darkblue!10} & 0.9581 \\
S-MVD-CL & 12.16                         & \textbf{0.9705} \cellcolor{darkblue!10}\\ \bottomrule
\end{tabular}
\label{tab:results-latent-space}
\end{table}

\begin{table*}[htbp]
\centering
\caption{Performance (mean\tiny{$\pm$std} \normalsize) of the various contrastive pre-training methods after fine-tuning across five folds using Dice Scores (\%), 95\% Hausdorff Distance ($\text{HD}_\text{95}$), and Average Surface Distance (ASD). For Dice Scores, we report the overall performance (DSC), specifically for healthy patients ($\text{DSC}_\text{NF}$) and for patients with Pleural Effusion ($\text{DSC}_\text{PE}$). Overall best results are highlighted in \textbf{bold}. Best results for both \colorbox{darkgreen!10}{unsupervised} and \colorbox{darkblue!10}{supervised} methods are also highlighted. }
\label{tab:results-mains}

\begin{tabular}{ll ccccc}
\toprule
& Pre-training Method  & DSC ($\uparrow$) & $\text{DSC}_\text{NF} (\uparrow)$ & $\text{DSC}_\text{PE} (\uparrow)$ & $\text{HD}_\text{95} (\downarrow)$ & $\text{ASD} (\downarrow)$ \\ \midrule

& None & 87.05{\tiny$\pm$4.34} & 93.59{\tiny$\pm$1.28} & 84.76{\tiny$\pm$1.28} & 21.9{\tiny$\pm$8.0} & 5.39{\tiny$\pm$2.05} \\ \midrule

\multirow{4}{*}{\rotatebox{90}{{\footnotesize Unsupervised}}} & SimCLR~\cite{chenSimpleFrameworkContrastive2020} & 90.90{\tiny$\pm$2.94} & 96.71{\tiny$\pm$0.28} & 88.61{\tiny$\pm$4.15} & 11.9{\tiny$\pm$5.3} & 3.64{\tiny$\pm$1.63} \\

& VADeR~\cite{o.pinheiroUnsupervisedLearningDense2020} & 92.26{\tiny$\pm$2.66} & 97.22{\tiny$\pm$0.16} & 90.33{\tiny$\pm$3.68} & 10.4{\tiny$\pm$3.7} & 2.79{\tiny$\pm$1.14} \\

& DVD-CL (\textit{ours}) & 92.76{\tiny$\pm$2.23} \cellcolor{darkgreen!10} & 97.36{\tiny$\pm$0.14} \cellcolor{darkgreen!10} & 91.08{\tiny$\pm$3.18} \cellcolor{darkgreen!10} & 10.5{\tiny$\pm$3.6} & 2.86{\tiny$\pm$1.04} \\

& MVD-CL (\textit{ours}) & 92.27{\tiny$\pm$2.73} & 97.18{\tiny$\pm$0.25} & 90.36{\tiny$\pm$3.79} & 9.1{\tiny$\pm$4.9} \cellcolor{darkgreen!10} & 2.61{\tiny$\pm$1.16} \cellcolor{darkgreen!10} \\ \midrule

\multirow{4}{*}{\rotatebox{90}{{\footnotesize Supervised}}} & CheXMask & 93.65{\tiny$\pm$2.33} & \textbf{97.81{\tiny$\pm$0.20}} \cellcolor{darkblue!10} & 92.02{\tiny$\pm$3.23} & 7.3{\tiny$\pm$1.8} & 2.46{\tiny$\pm$1.26} \\

& SSDCL~\cite{huSemisupervisedContrastiveLearning2021} & 92.91{\tiny$\pm$2.77} & 97.58{\tiny$\pm$0.17} & 91.10{\tiny$\pm$3.84} & 15.3{\tiny$\pm$5.0} & 3.31{\tiny$\pm$1.20} \\

& S-DVD-CL (\textit{ours}) & 93.15{\tiny$\pm$2.61} & 97.62{\tiny$\pm$0.11} & 91.43{\tiny$\pm$3.65} & 8.0{\tiny$\pm$3.9} & 2.96{\tiny$\pm$1.97} \\

& S-MVD-CL (\textit{ours}) & \textbf{93.93{\tiny$\pm$1.92}} \cellcolor{darkblue!10} & 97.75{\tiny$\pm$0.09} & \textbf{92.43{\tiny$\pm$2.67}} \cellcolor{darkblue!10} & \textbf{7.3{\tiny$\pm$3.4}} \cellcolor{darkblue!10} & \textbf{2.05{\tiny$\pm$0.74}} \cellcolor{darkblue!10} \\ \bottomrule

\end{tabular}%
\end{table*}

\subsection{Latent Space Analysis}

We evaluate the learned embedding using the proposed methods, both qualitatively (with CHRO-Maps) and quantitatively. For quantitative results, the output embeddings are clustered into three clusters using K-means clustering, and the purity of each cluster was calculated using the CheXmask labels. Purity is calculated as a ratio of the majority-labeled pixel in a cluster over the total number of pixels in the cluster~\cite{chen1998speaker}. A higher value of purity represents better clustering. In addition, utilizing the same CheXmask labels, we calculate the ratio of inter-class distance and intra-class standard deviation, which should increase as the model clusters pixels from the same class together~\cite{calinski1974dendrite}.
From both Fig.~\ref{fig:chro-maps} and Table~\ref{tab:results-latent-space}, we observe that in unsupervised variants, for DVD-CL, no sharp clusters are visible; however, each lung occupies a distinct region of the latent space. The CHRO-maps further reveal that the model encodes pixels according to their spatial relationship to the spine, indicating a degree of positional awareness in the learned representations. In contrast, MVD-CL fails to learn meaningful semantic structure, as evidenced by its poor clustering performance in Table~\ref{tab:results-latent-space}. We also observe that the supervised variants (S-DVD-CL and S-MVD-CL) learn well-defined clusters corresponding to the three anatomical classes. In the CHRO-maps (Fig.~\ref{fig:chro-maps}), the boundaries between the lungs and the background are also assigned distinct colors, suggesting that the model captures inter-class boundary information in its latent representations. In Table~\ref{tab:results-latent-space}, we observe that both supervised variants achieve high clustering purity and distance to variance ratio.


\subsection{Fine-tuning Evaluation}

\begin{figure}[t]
    \centering
    \includegraphics[width=1.0\linewidth]{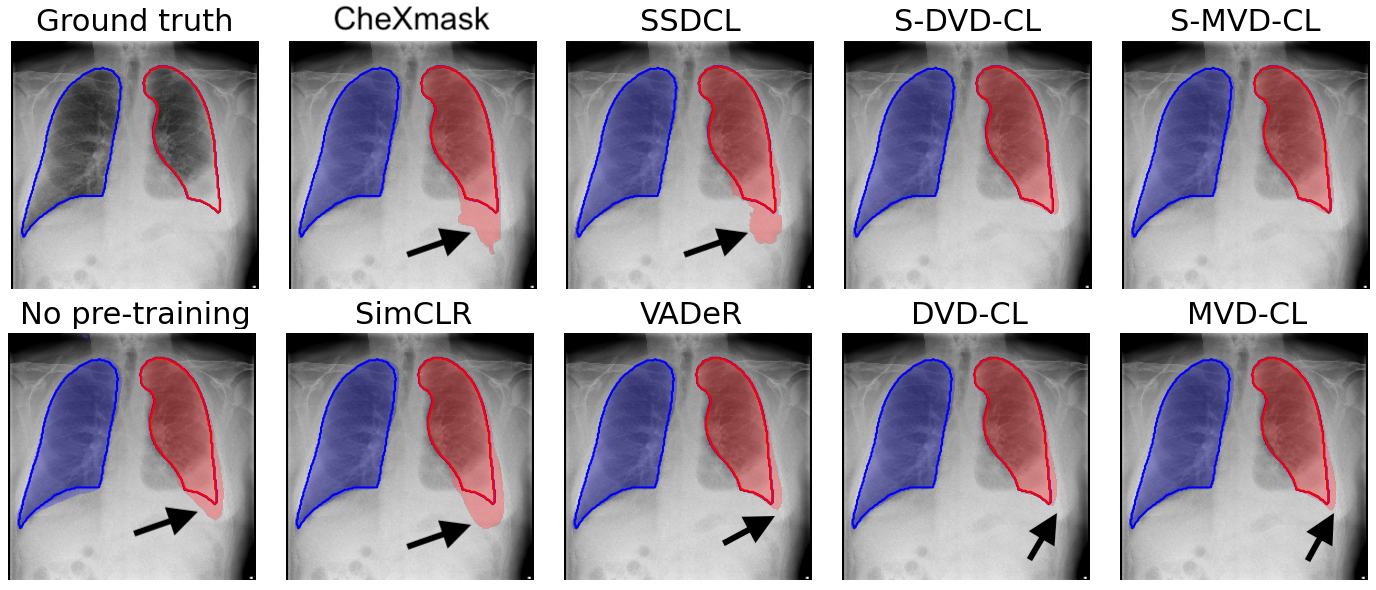}
    \caption{Qualitative results for lung segmentation using various pre-training methods on PE image. Supervised methods (top row) show better performance compared to unsupervised methods (bottom row). We also observe that the proposed methods outperform their counterpart baselines. }
    \label{fig:quali-results}
\end{figure}

We fine-tune all pre-trained models on the manually annotated PadChest subset~\cite{mehtaCFSegCounterfactualsMeet2025} using a five-fold cross-validation approach. As shown in Table~\ref{tab:results-mains} and Fig.~\ref{fig:quali-results}, all proposed pre-training methods outperform their respective baselines, indicating that pixel-level counterfactual contrastive pre-training produces more transferable representations for segmentation. Among the unsupervised methods, DVD-CL achieves the best overall performance, outperforming SimCLR and VADeR. The use of counterfactual augmentations further improves generalisation as reflected by the lower variance across folds. Both supervised variants, S-DVD-CL and S-MVD-CL, further improve performance, demonstrating the benefit of incorporating label information when available. Notably, S-MVD-CL surpasses the \textit{silver-standard} CheXMask pre-trained model, achieving an average DSC of $\sim$94\% on challenging manually annotated samples. This highlights that dense, multi-view contrastive learning with pixel-level supervision can yield representations that transfer well to difficult samples, even from \textit{silver-standard} pre-training data.





\section{Conclusion}
\label{sec:conclusion}

In this work, we introduced a family of pixel-level counterfactual contrastive learning methods designed for medical image segmentation. By combining counterfactual image generation with dense contrastive objectives, our approach enables the learning of representations that are both spatially consistent and invariant to confounding imaging factors such as scanner type or disease presence. Through extensive experiments, we demonstrated that our unsupervised methods outperform existing self-supervised baselines, while the supervised variants further improve performance and robustness. Notably, both S-DVD-CL and S-MVD-CL outperform supervised segmentation pre-training when given access to identical data. The proposed CHRO-map visualisation also provides an intuitive way to interpret pixel embeddings and assess representation quality.
Overall, our results demonstrate that integrating counterfactual reasoning into dense contrastive learning presents a promising approach to developing more interpretable and robust medical segmentation models. However, our counterfactual generation approach assumes the causal graph is known and that counterfactuals are identifiable from observed data. These assumptions may not hold in practice, affecting the causal validity of our estimates~\cite{ribeiro2025counterfactualidentifiabilitydynamicoptimal}. Future work may explore alternative counterfactual generation methods to address these limitations, as well as scaling the framework to 3D modalities or semi-supervised settings.


\section{Compliance with ethical standards}
\label{sec:ethics}
This study uses secondary, fully anonymised data which is publicly available and is exempt from ethical approval.

\section{Acknowledgments}
\label{sec:acknowledgments}
This project was partially supported by the Royal Academy of Engineering (Kheiron/RAEng Research Chair), the UKRI AI programme, and the EPSRC, for CHAI - EPSRC Causality in Healthcare AI Hub (grant no. EP/Y028856/1), and the European Union’s Horizon Europe research and innovation programme under grant agreement 101080302.

\bibliographystyle{IEEEbib}
\bibliography{strings,refs}

\end{document}